\documentclass[11pt,letterpaper]{article}
\usepackage{hyperref}
\usepackage{emnlp2016}
\usepackage{times}
\usepackage{latexsym}
\usepackage{amssymb}
\usepackage{amsmath}
\usepackage{graphicx}
\usepackage{tabularx}
\usepackage{color}

\AtBeginShipout{%
  \ifnum\value{page}>1 %
    \typeout{* Additional boxing of page `\thepage'}%
    \setbox\AtBeginShipoutBox=\hbox{\copy\AtBeginShipoutBox}%
  \fi
}

\newlength\replength
\newcommand\repfrac{.33}

\newcommand\rulewidth{.6pt}
\newcommand\tdashfill[1][\repfrac]{\cleaders\hbox to \replength{%
  \smash{\rule[\arraystretch\ht\strutbox]{\repfrac\replength}{\rulewidth}}}\hfill}
\newcommand\tabdashline{%
  \makebox[0pt][r]{\makebox[\tabcolsep]{\tdashfill\hfil}}\tdashfill\hfil%
  \makebox[0pt][l]{\makebox[\tabcolsep]{\tdashfill\hfil}}%
  \\[-\arraystretch\dimexpr\ht\strutbox+\dp\strutbox\relax]%
}
\newcommand\tdotfill[1][\repfrac]{\cleaders\hbox to \replength{%
  \smash{\raisebox{\arraystretch\dimexpr\ht\strutbox-.1ex\relax}{.}}}\hfill}

\emnlpfinalcopy

\def\emnlppaperid{1119}

\newcommand{\full}{{\mbox{\small {\sc Full}}}}
\newcommand{\human}{{\mbox{\small {\sc Human}}}}
\newcommand{\syn}{{\mbox{\small {\sc -Syn}}}}
\newcommand{\sem}{{\mbox{\small {\sc -Sem}}}}
\newcommand{\swDev}{{\mbox{\small {\sc Star$_{dev}$}}}}
\newcommand{\swTest}{{\mbox{\small {\sc Star$_{test}$}}}}
\newcommand{\cartoon}{{\mbox{\small {\sc Cartoon}}}}

\title{A Theme-Rewriting Approach for Generating Algebra Word Problems}
 \author{Rik Koncel-Kedziorski \\ {\bf Ioannis Konstas} \\{\bf Luke Zettlemoyer} \\ {\bf Hannaneh Hajishirzi} \\
         University of Washington \\ kedzior@uw.edu, \{ikonstas, lsz, hannaneh\}@cs.washington.edu}
\date{}

\begin{document}

\maketitle
\begin{abstract}
Texts present coherent stories that have a particular theme or overall setting, for example science fiction or western. In this paper, we present a text generation method called {\it rewriting} that edits existing human-authored narratives to change their theme without changing the underlying story. We apply the approach to math word problems, where it might help students stay more engaged by quickly transforming all of their homework assignments to the theme of their favorite movie without changing the math concepts that are being taught. 
Our rewriting method uses a two-stage decoding process, which proposes new words from the target theme and scores the resulting stories according to a number of factors defining aspects of syntactic, semantic, and thematic coherence.
Experiments demonstrate that the final stories typically represent the new theme well while still testing the original math concepts, outperforming a number of baselines. 
We also release a new dataset of human-authored rewrites of math word problems in several themes. 
%Generating a thematically diverse set of coherent stories is important in different domains from entertainment to education. We present a text generation method called {\it rewriting} which leverages the coherence of human-authored narratives to achieve novel and coherent stories in a new theme. Our focus is to generate  the most thematic math word problem that is syntactically and semantically coherent given a human-authored problem and the new theme. Our method uses a two-stage decoding process.  First, it generates candidate problems from the space of all possible problems in a new theme. Second, it scores the generated stories by applying local semantic, syntactic, as well as inter- and intra-sentential discourse scores. We evaluate the application of this method in generating  personalized math word problems for students, showing xxx. In addition to the {\it rewriting} method, we release a dataset of over 500 human-authored rewrites of math word problems in several themes used for training and evaluation. 
\end{abstract}

\section{Introduction}

%Storytelling is a complex activity that is at the same time fundamental to the human experience and understanding of the world [cite psychologists]. 
Storytelling is the complex activity of expressing a plot, its events and participants in words meaningful to an audience.  
%A storyteller must posses supreme command of syntactic and discourse structures as well as semantic knowledge of the plot in order to construct a coherent narrative.
Automatic storytelling systems can be used for customized sport commentaries, enriching video games with personalized or dynamic plot-lines~\cite{barros2007planning}, or  providing customized learning materials which meet each individual student's needs and interests~\cite{bartlett2004expanding}. 
In this paper, we focus on generating narrative-style math word problems (Figure~\ref{fig:teaser}) and demonstrate that it is possible to design an algorithm that can automatically change the overall theme of a text without changing its underlying story, for example to create more engaging homework that is in the theme of a student's favorite movie.

%It is, however, challenging to build an automated storyteller that produces entire new text, and it is worth considering the extent to which existing stories can be automatically editing or otherwise modifying existing work.
%that copes with syntactic, semantic, and discourse understanding required to effect a well-wrought story.

%Moreover, intricate semantic knowledge of the plot, its events and participants is necessary to render a story meaningful to its audience.

%Yet the value of an automated storytelling agent is undeniable. 
%Story generation can be of use in entertainment, for example by enriching video games with personalized or dynamic plot-lines \cite{barros2007planning}. 
%It can also be of great use in the domain of educational technology.
%Here, an storytelling AI could assist overworked educators\footnote{\cite{bartlett2004expanding}} by providing customized learning materials which meet each individual student's needs and interests. 

\begin{small}
\begin{figure}[t]
\begin{centering}
  \begin{tabularx}{\columnwidth}{|X|}
    \hline 
%\small Chris has been saving his allowance to buy a new pair of soccer cleats and a ball. His grandmother gave Chris \$25 for his birthday. His aunt and uncle gave Chris \$20 and his parents gave him \$75. Now Chris had \$279. How much money did Chris have before his birthday? \\ \hline
%\small \texttt{Ben Kenobi has been saving his gold to buy a new pair of station rooms and a starship. His friend gave ben kenobi \$25 for his unit. His aunt and uncle gave ben kenobi \$20 and his officers gave him \$75. Now Ben Kenobi had \$279. How much money did Ben Kenobi have before his unit?}
\small {\bf Jim walked 0.2 of a mile from school to David's house and 0.7 of a mile from David's house to his own house. How many miles did Jim walk in all?} \\
\hline
{\bf Star Wars} \\ \small Uncle Owen walked 0.2 of a mile from hangar to Luke Skywalker's room and 0.7 of a mile from Luke Skywalker's room to his own room. How many miles did Uncle Owen walk in all? \\
\hline
{\bf Cartoon } \\  \small Finn squished 0.2 of a mile from cupboard to Melissa's dock and 0.7 of a mile from Melissa's dock to his own dock. How many miles did Finn squish in all?\\
\hline
{\bf Western} \\
\small Duane strolled 0.2 of a mile from barn to Madeline's camp and 0.7 of a mile from Madeline's camp to his own camp. How many miles did Duane stroll in all?\\
\hline

\end{tabularx}
\caption{An example story and rewrites in 3 themes.}
\label{fig:teaser}
\end{centering}
\end{figure}
\end{small}

%The utility of a story in an educational setting is stronly determined by its coherence. 
%In this paper, we focus on generating narrative-style math word problems (Figure~\ref{xx}). 
A math word problem is a coherent story that provides the student with good clues to the correct mathematical operations between the numerical quantities described therein.
%, as well as the correct order of their application.
However, the particular {\it theme} of a problem, whether it be about collecting apples or traveling distances through space, can vary significantly so long as the correlation between the story and underlying equation is maintained. 
Students' success at solving a word problem is tied to their interest in the problem's theme~\cite{renninger2002individual}, and
personalizing word problems increases student understanding, engagement, and performance in the problem solving process~ \cite{hart1996effect,davis1991role}.

Motivated by this need for thematically diverse, highly coherent stories, we address the problem of {\it story rewriting}, or transforming human-authored stories into novel, coherent stories in a new theme. %by rewriting  human authored text into novel, coherent stories in a new theme. %We introduce a method for story rewriting that allows us to leverage the coherence of human authored text in the production of novel stories. 
Rather than synthesizing first a story plot \cite{mcintyre2009learning,mcintyre2010plot} or script \cite{chambers2009unsupervised,pichotta:aaai14,granroth:aaai16} from scratch, we instead begin from an existing story and iteratively edit it towards a thematically novel but --most crucially-- semantically compatible story. This approach allows us to reuse much, but not all, of the syntactic and semantic structure of the original text, resulting in the creation of more coherent and solvable math word problems. 

We define a theme to be a collection of reference texts, such as a movie script or series of books. 
Given a theme, the {\it rewrite} algorithm constructs new texts by substituting thematically appropriate words and phrases, as measured with automatic metrics over the theme text collection, for parts of the original texts. This process optimizes for a number of metrics of overall text quality, including syntactic, semantics, and discourse scores. It uses no hand crafted templates and requires no theme-specific tuning data, making it easy to apply for new themes in practice. Tables \ref{examples_star_wars}--\ref{examples_western} show example stories generated from the rewrite system.

To evaluate performance, we collected a corpus of 450 rewrites of math word problems in Star Wars and Children's Cartoon themes via crowdsourcing.\footnote{Data and code available at \url{https://gitlab.cs.washington.edu/kedzior/Rewriter/}.}
Experiments with automated metrics and human evaluations demonstrate that the approach described here outperforms a number of baselines and can produce solvable problems in multiple different themes, even with no in-domain tuning. 

%We make this data, as well as a demo of our system which can rewrite arbitrary stories in a new theme, available as a resource for other researchers and for educators to use in the classroom. Experiments demonstrate that xxx

% The contributions of this work are threefold: First, we build the first automated system to address the new problem of story rewriting, making the software and data available for future use. Second, we formalize the problem of story rewriting as learning to narrate coherent stories in a new theme by optimizing for syntactic, semantic, and discourse structure of the story. Third, we introduce a method for deriving new themes and adapting stories in different themes.

\section{Related Work}
Our approach is related to the previous work in story generation (e.g.,~\newcite{mcintyre2010plot}) and sentence rewriting (e.g., text simplification~\cite{weiXu16}), as reviewed in this section. It has three major differences from all these approaches: First, we focus on multi-sentence stories where preserving the coherence, discourse relations,  and solvability is essential. Previous  work  mainly focuses on rewriting single sentences. Second, we build a  theme from a text corpus and show how the stories can be adapted to new themes. Third, our method leverages the human-authored story to capture the semantic skeleton and the plot of the current story, rather than synthesizing the story plot.  To our knowledge, we are the first to introduce a text rewriting formulation for story generation.  

Story generation has been of long interest to AI researchers~\cite{meehan1976metanovel,lebowitz1987planning,turner1993minstrel,liu2002makebelieve,nasrin}. %, with early methods focusing on the planning aspect of storytelling \cite{meehan1976metanovel,lebowitz1987planning,turner1993minstrel} or incorporation of commonsense knowledge for collaborative storytelling \cite{liu2002makebelieve}. %More recent work use crowd-sourced plot graphs~\cite{Li13} or structured representations~\cite{mcintyre2010plot} to generate stories.
Recent methods in story generation first  synthesize candidate plots  for a story and then compile those plots into text.  \newcite{Li13} use crowdsourcing to build plot graphs.  McIntyre and Lapata  \shortcite{mcintyre2009learning,mcintyre2010plot} 
address story generation through the automatic deduction and reassembly of scripts \cite{schank1977scripts}, or structured representations of events and their participants, and causal relationships involved.
Leveraging the automatic script learning methods of \newcite{chambers2009unsupervised}, \newcite{mcintyre2010plot}  learn candidate entity-centered plot graphs, or possible events involving the entity and an ordering between these events, with the use of a genetic algorithm.
%In , they use a genetic algorithm to mix and mutate plot graphs to produce (ideally) stronger plot candidates. 
Then plots are compiled into stories through the use of a rule-based text surface realizer \cite{lavoie1997fast} and reranked using a language model.
% and the most probable story is selected by a language model. 

%While contributing to variety in the production of new plots, event co-occurrence
%as a content planner component oversimplifies the definition of \textit{plot},
%reducing to less coherent or more mundane stories compared to human-authored. More crucially the stochasticity of the genetic algorithm could inadvertently harm the semantic soundness of the plot, a factor crucial for the domain of math word problems and their potential solvability.
%Responding to this outcome, the current work assumes the original human-authored plot and consistently deviates from it subject to semantic, syntactic, and discourse coherence constraints.
% While this method is of value in the production of new plots, it must significantly oversimplify the definition of "plot" to do so. 
% Plot is reduced to event co-occurrence, and the resulting generated plots are drastically less coherent and less interesting than human-authored stories. 
% Responding to this outcome, the current work focuses on deviating from established (i.e. human-authored) plots rather than attempting to take on the difficult task of plot generation. 

%Work on automatically generating math word problems tailored to a student's interest includes \newcite{polozov2015personalized}. 
% Their method makes use of Answer Set Programming to satisfy a collection of pedagogical and narrative requirements. 
\newcite{polozov2015personalized} automatically generate math word problems tailored to a student's interest using Answer Set Programming to satisfy a collection of pedagogical and narrative requirements. %Starting from the equation they formulate a content plan in the form of a logical graph using events and entities from an ontology. Semantic  coherence based on predefined \textit{discourse tropes} is applied as an extra set of constraints in the graph; the resulting text is generated using hand-crafted templates.
% First, they generate an equation based on the pedagogical requirements. 
% From this equation, they can then generate a plot in the form of a logical graph using events and entities from an ontology.
% To ensure these plots are coherent, they formalize the notion of a {\it discourse trope} as a constraint on the form of logical graphs and force their graphs to obey these constraints. 
% Finally, they use a template based text generator for surface realization of the graph. 
This method naturally produces highly coherent, personalized story problems that meet pedagogical requirements, at the expense of building the thematic ontologies and discourse constraints by hand.\footnote{According to \newcite{polozov2015personalized} building small thematic ontologies of types, relations, and discourse tropes (100-200 entries) for each of only 3 literary settings took 1-2 person months.}

Additionally, there is related work in text simplification~\cite{wubben2012sentence,kauchak13,Zhu2010,Vanderwende,Woodsend:Lapata:11b,hwang2015aligning}, sentence compression~\cite{Filippova08,rush2015neural}, and paraphrasing~\cite{ganitkevitch2013ppdb,Chen13,Ganitkevitch11}. All these tasks are focused on rewriting sentences under a predefined set of constraints, such as simplicity.  Different rule-based and data-driven approaches are introduced by \newcite{PetersenOst07}, \newcite{vickrey08}, and \newcite{Siddharthan04}. Most data-driven approaches take advantage of machine translation techniques, use source-target sentence pairs, and learn rewrite operations~\cite{Yatskar10a,Woodsend:Lapata:11a}, or use additional external paraphrasing resources~\cite{weiXu16}. % Recently,    adapt machine translation techniques\newcite{weiXu16} for text simplification to find the most simplified text  by optimizing a set of metrics. To compensate for low quality simplification data, their method uses a small set of manual  simplification sentence pairs and a large scale paraphrase dataset. 

Finally, this work is related to those on automatically solving math word problems.
Specific topics include number word problems~\cite{shi15automatically}, logic puzzle problems~\cite{mitralearning}, arithmetic word problems~\cite{hosseini2014learning,roy2015solving}, algebra word problems~\cite{kushman2014learning,zhou2015learn,koncel2015parsing,roy2016equations}, and geometry word problems~\cite{seo2015solving,seo2014diagram}. 
Several datasets of word problems are available~\cite{koncel:naacl16,huangwell}, though none address the need for thematic text.

%Similar to \newcite{weiXu16}, we take advantage of tuning metrics on a set of human-authored story rewrites and use external resources to facilitate the rewriting.  

\section{Problem Formulation}

% The high-level goal of our study is to rewrite a short story in a given theme to a different, possibly distant theme while preserving the narrative coherence or readability of the text.
%  In this paper, we focus on generating novel, solvable math word problems in a given theme. To that purpose, we use a dataset of available human-authored math word problems and rewrite each problem in a new theme (Figure~\ref{xx}).  
 
% The generated math word problems should be readable, coherent, and solvable. These properties of a story problem derive in part from the relationships between syntactic structures used in the problem, as well as the semantic relationships between different elements of the problem.  For a specific story, we show it is possible to leverage  syntactic and semantic relationships encoded in the original text to facilitate the generation of a new, coherent story in a different theme. 
% Figure~\ref{arrows} gives an overview of the information from the original story that is transferred to rewrites.

Our system takes as input a story $s$ and a theme $t$, and outputs the best rewrite $s^{*}$ from generated candidates $S$. 

%A theme $t$ is defined as a collection of documents specific to a particular genre (e.g., movie scripts, TV show, and novels). 
A theme $t$ is defined as a textual corpus that describes a topic or a domain. 
This is an intentionally broad definition that allows a variety of textual resources to serve as themes.
%\sout{Here, we use a flexible notion of a theme and allow a variety of text to serve as unique themes. }
For example, the collection of all Science Fiction stories from the Project Gutenberg can be a theme, or the script of a single movie, or a sampling of fan fiction from the Internet. 
This flexibility adds to the utility of our work, as varying amounts of thematic text may be available.

The generated candidate $s^{*}$ is the most {\it thematically} fit problem that is {\it syntactically} and {\it semantically} coherent given the original problem $s$ and the new theme $t$.  We represent a story in terms of the words it contains, so that $s = \{w_1, w_2, \ldots\ w_n\}$ and $|s|=n$. The new story $s^{\prime}$ is defined as:
\[
    s^{\prime} = \bigl\{f(w_1), f(w_2), \ldots\ f(w_n)\bigr\}
\]
where the function $f(w) : \mathcal{V}_o \rightarrow \mathcal{V}_t^K \cup \varnothing$, rewrites a word from the vocabulary of the original problem $\mathcal{V}_o$ to either a word, a trivial noun compound of length $K$ (e.g., multi-word named entity) from the vocabulary of the the thematic vocabulary $\mathcal{V}_t$, or reduces to the empty symbol, i.e., omits the input word entirely; hence the length of $s^{\prime}$ can differ from that of the original problem.

%Similarly, the new story $s'$ is a set of words $s'=\{w'_1,\ldots, w'_k\}$, where $w'_i=f(w_i)$  is an adaptation of the word $w_i$ in the original story.

\begin{figure}[t]
    \centering
    \includegraphics[scale=0.35]{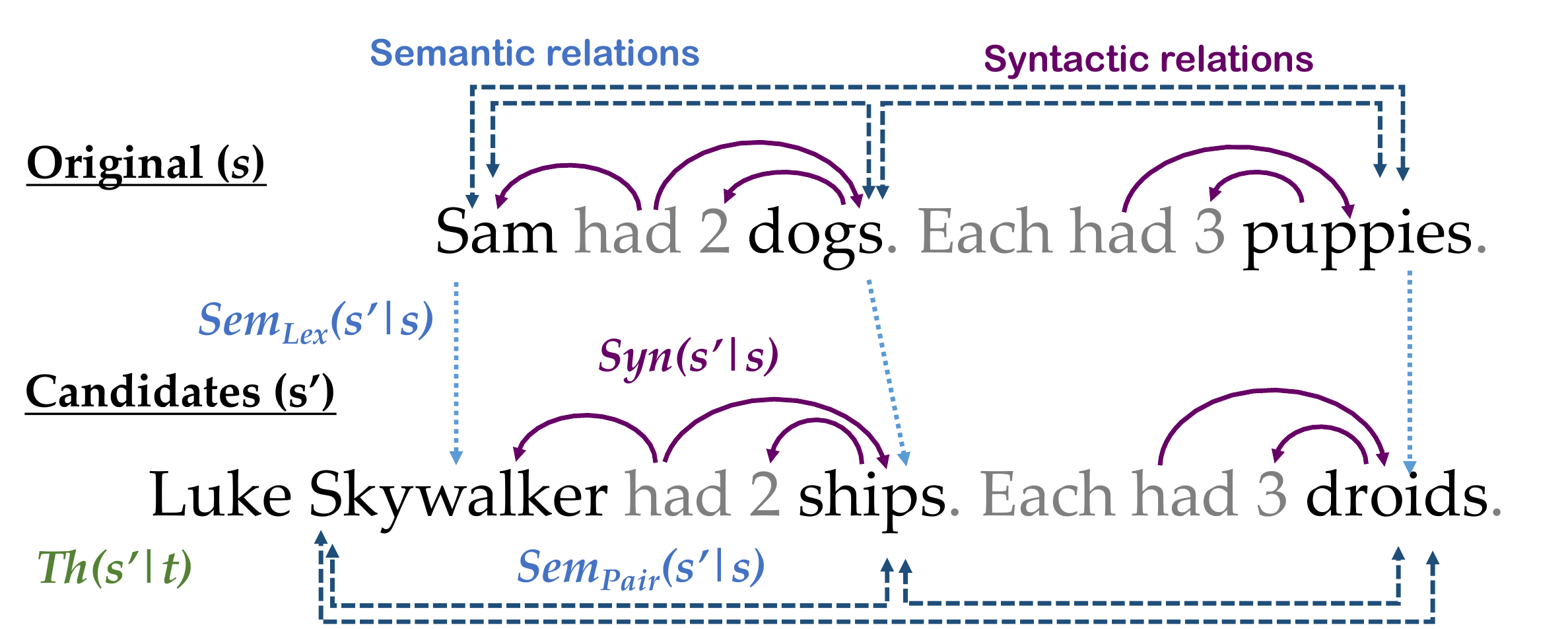}
    \caption{\small An overview of our method for scoring a candidate story $s'$ given a human-authored story $s$ and a theme $t$. Syn$(s'|s)$: compatibility  of syntactic relations (purple arrows),  Sem$_{pair}(s'|s)$: coherence of semantic relations (blue arrows),  $\mbox{Sem}_{Lex}(s'|s)$: semantic mapping of individual words, and  $Th(s'|s,t)$: thematicity.}
    \label{fig:overview}
\end{figure}

Formally, our goal is to select the candidate \mbox{$s^{\prime} \in S$} by maximizing a scoring function $\mathcal{R}$ over thematic, syntactic and semantic constraints, subject to a set of parameters $\theta$:
\begin{equation}
  s^*=\arg\max_{s^{\prime} \in S} \mathcal{R} (s^{\prime} | s,t; \theta)
\end{equation}

In order to find the best story $s^*$, our problem reduces to generating candidate stories $s'$ from the space of possible rewrites of the human-authored story $s$ in a new theme $t$ (Section~\ref{sec:decoding}). Since there are exponentially many rewrites, we follow a two-stage decoding approach: first we identify only the content words $w_i$ in the input problem, and provide for each a list of the top-${k}$ most {\it salient} thematic candidate words and trivial noun compounds. We then search the space by progressively introducing more rewrites in the beam, and scoring them according to $\mathcal{R}$ (Section~\ref{sec:scoring}). Figure~\ref{fig:overview} shows the overview of the scoring function for a candidate sentence $s'$.  
%Our method then uses a scoring function to evaluate the score of $s'$ given the original story $s$ and the theme $t$ (Section~\ref{x}). The scoring function incorporates local semantic, syntactic, as well as inter- and intra-sentential discourse scores at each step. %The final beam is reranked using a language model to ensure surface grammaticality. 

% In order to tune the parameters of the scoring function, we collect a corpus of human rewrites of each story $s$ and then maximize the similarity between generated rewrites and human rewrites. 

%The complexity of these relationships in the abstract is beyond what can be computed by present text generation techniques. 

%The rewriting task can be considered abstractly in two parts. 
%On the one hand, we are inducing a story template from the original story to aid in the rewriting process.
%This template consists of information-heavy syntactic structures and semantic relationships that we wish to preserve from original.
%We assert that the complexity of these structures in narrative is beyond what can be computed by present methods, and we choose to obtain this information instead from the human-authors of the original story.
%At the same time, we are automatically inducing from the theme text a story-template-filler, allowing us to immediately fill the template created from the original with theme-specific entities, events, and ideas. 

%%\begin{figure}[t]
%    \centering
%    \includegraphics[scale=0.35]{decoding2.pdf}
%    \caption{Decoding process.}
%    \label{fig:overview}
%\end{figure}

\section{Scoring Stories}\label{sec:scoring}%Informally, themes are topics or domains of interest.
The scoring function $\mathcal{R}$ %computes the score of a candidate generated story $s'$ given the original story $s$ and a theme $t$.
decomposes into three components, capturing aspects of syntactic compatibility, semantic coherence, and thematicity: 
\begin{align}\label{eq:scoring_function}
  \mathcal{R}(s^{\prime} | s,t;\theta) =  & \alpha \times \textrm{Sem}(s^{\prime} | s) \nonumber \\
                                 &+ \beta \times \textrm{Syn}(s^{\prime} | s) \\
                                 &+ \gamma \times \textrm{Th}(s^{\prime} | s,t) \nonumber
\end{align}
The syntactic (Syn) and semantic (Sem) coherence components measure the coherence of the words in the new story $s^\prime$, as well as their compatibility to the syntactic and semantic relations in the original story $s$. On the other hand, thematicity (Th) scores the relevance and importance of words in the new story with respect to theme $t$. 

We describe each of these components and the decoding process in the following sections. 

%  \begin{figure}[t]
%  \begin{center}
%    \includegraphics[width=0.45\textwidth]{arrows}
%  \end{center}
%  \caption{Information from $s$ used in $s^{\prime}$. Here, $w_i$ and $w_j$ are the $i^{\textrm{th}}$ and $j^{\textrm{th}}$ words in the original story $s$, and $w^{\prime}_i$ and $w^{\prime}_j$ are the $i^{\textrm{th}}$ and $j^{\textrm{th}}$ words in a candidate rewrite $s^{\prime}$.}
%  \label{arrows}
%\end{figure}

\subsection{Thematicity}

Recall that a theme $t$ is defined as a collection of documents that share a common topic, such as books in the science fiction genre, or scripts of horror movies. We define thematicity of a word $w^{\prime}$ as the measure of {\it salience}, or how discriminative that word is to a given theme.\footnote{We will be interchangeably referring to $w^\prime$ as either the word or the head of the multi-word noun compound that rewrites the equivalent word $w$ in the original problem.} 
For example, \textit{robot} and \textit{spaceship} are expected to be highly thematic with respect to Star Wars.
In our setting we extend this definition to a candidate problem $s^{\prime}$ given $s$ and $t$ as:
\begin{equation}
\textrm{Th}(s^{\prime} | s, t) =\sum_{i}^{|s|} Sal(w_i',t)%+\textrm{LM}_t (s^{\prime})
\end{equation}
where $w^{\prime}_i$ is a word from the candidate problem, and $Sal$ is its salience score with respect to the theme. In the context of this work we argue that the thematic adaptation of the content words, i.e., nouns, verbs, named entities, and adjectives, plays the most important role in forming a new thematic problem. Therefore, we define their salience (except named entities) based on their tf-idf score over the theme $t$, and set it to zero for function words. Since named entities have relatively low frequencies in the theme corpus we set their salience to $1-\frac{1}{c(w_i^{\prime})}$, where $c(w_i^{\prime})$ is the number of times $w_i^{\prime}$ occurs in the theme. In the example story in Figure~\ref{fig:overview} the thematicity score is derived as $Sal(\mbox{Luke Skywalker})+Sal(\mbox{ships})+Sal(\mbox{droids})$.

\subsection{Syntactic compatibility}
This work offers a new method for syntactic and discourse coherence based on preserving human-authored syntactic structure in generated text (hence our use of the term {\it rewriting}). The syntactic constructs in a document play a distinctive role in maintaining cohesion across sentences.
We consider the human-authored syntax of the original story $s$ as gold standard, and use it to score a candidate problem $s^{\prime}$ 
%based on the syntactic structure of the original story $s$ 
by considering how well the syntactic relations of $s$ apply to $s^{\prime}$.

% Note that, while discourse coherence is comprised of many features, including semantic features to be discussed later, it is partly encoded in syntactic structure [cite who?].
%We observe that preserving syntax is particularly helpful in maintaining discourse coherence in the generated stories. 
% We consider the human-authored syntax of the original story $s$ as a gold standard for the structuring of the information necessary for coherently expressing a similar story to $s$. 
% We score a candidate problem $s^{\prime}$ based on the syntactic structure of the original story $s$ by considering how well the syntactic relations of $s$ apply to $s^{\prime}$.

Formally, given a dependency triple $(w_i,w_j,l)$ from a parse of a sentence in $s$, we compute the likelihood for the corresponding triple $(w^{\prime}_i,w^{\prime}_j,l)$ for $w^{\prime}_i,w^{\prime}_j$ in $s^{\prime}$.  
We define the syntactic score for all sentences in $s^{\prime}$ as: 
% In particular, it scores $s^{\prime}$ according to how well the words of $s^{\prime}$ work when configured according to the dependency relations of $s$.
% Therefore The syntactic compatibility score for $s^{\prime}$ is then the sum of these likelihoods:
\begin{equation}\label{eq:syn}
 \hspace{-.1cm} \textrm{Syn}(s^{\prime} | s) =  \sum_{i,j,l | (w_i,w_j,l) \in \textrm{Dep}(s)}\mathcal{L}_{Dep}(w^{\prime}_i,w^{\prime}_j,l)
\end{equation}
where $Dep(s)$ are the dependency parse trees for all sentences in $s$; $\mathcal{L}_{Dep}$ is a 3-gram language model over dependency triples which gives the likelihood of an arc label $l$ being used between a pair of words $(w_i',w_j')$. For example in Figure~\ref{fig:overview}, the syntactic compatibility score includes dependency likelihoods of $\mathcal{L}_{Dep}(\mbox{ship}, \mbox{2},\mbox{{\it num}})$, $\mathcal{L}_{Dep}(\mbox{had},\mbox{ship},\mbox{{\it dobj}})$. 
% This model computes the likelihood that an arc label $l$ is used between a pair of words $(w_i',w_j')$ in a corpus. % connected with an arc lable $l$ and returns a score for the syntactic configuration. 

Therefore, the Syn function prefers stories $s'$ that (a) have similar dependency structure to the original story $s$ and (b) make use of a common syntactic configuration. %, while the selection of words prefer  goal of the $Syn$ function is to (a) keep the syntactic structure of the  give weight to those $s^{\prime}$ which make use of common syntactic configurations. 
%Since the dependency labels are determined by $s$, however, it scores $s^{\prime}$ according to how well the words of $s^{\prime}$ work when configured according to the syntax of $s$.

\subsection{Semantic Coherence}
%Another key component of the discourse coherence of a story is its semantics.
The semantic coherence component expresses how well a candidate $s^{\prime}$ rewrites individual words and realizes the semantic relationships that exist in the human-authored story $s$. Ideally, we would like to preserve enough of the semantics of $s$ in order to produce a coherent story $s^{\prime}$, yet we are populating $s^{\prime}$ with words taken from an unrelated theme. Therefore, we model the semantics of a story $s^\prime$ in terms of the lexical semantics contributed by individual words as well as semantic relationships that exist between its elements. Note that the relationships can cross the sentence boundaries, promoting discourse coherence.% of $s$. 

%We take advantage of the available lexico-semantic resources such as WordNet and Vector-space word embeddings to  
We decompose semantic relations in a story into a set of local, lexical relationships between pairs of words. Specifically, we
consider semantic relations for noun-noun and verb-verb pairs as provided by WordNet~\cite{miller1995wordnet}. Since some relations
are not directly outlined in these resources (e.g., the selectional preferences of nouns with regard to their adjectival modifiers), 
we also consider the word-embedding similarity between words.  For example in Figure~\ref{fig:overview} the semantic relationships are denoted with blue arrows between pairs of content words in the story (e.g., \{Sam, dogs\}, \{dogs, puppies\}, etc). 
% Lexico-semantic resources such as WordNet provide some of these semantic relations between words
% To account for these and  lexical correspondences, we make use of a vector-space embedding of words as a semantic resource. 
%Vector-space word embeddings are alleged to capture interesting semantic regularities \cite{mikolov2013linguistic}, which motivates their use in our task.

% More formally, the semantic coherence of the  new story $s$ with respect to the original story $s'$ has two components:  the compatibility between individual words $LexMap(w_i,w_i')$ and the compatibility between local semantic relations of words $PairMap(\{w_i,w_j\}, \{w_i',w_j'\})$.

More formally, we define the semantic coherence of $s^\prime$ with respect to $s$ as: 
\begin{align}
  \textrm{Sem}(s'|s) =  & \sum_{i}^{|s'|} \textrm{Sem}_{Lex}(w_i, w_i')  \\
                                 &\hspace{-.7cm}+  \sum_{i,j \in CW} \textrm{Sem}_{Pair}(\{w_i,w_j\}, \{w_i',w_j'\}) \nonumber
\end{align}
where $CW$ is the set of pairs of indices of content words (nouns, verbs, adjectives, and named entities) from $s$. We focus on the content words of the original problem, as they carry most of the semantic information. $\mbox{Sem}_{Lex}$ and $\mbox{Sem}_{Pair}$ functions are semantic adaptation scores for individual words and semantic relations respectively, described below.

Semantic Compatibility between words ($\mbox{Sem}_{Lex}$) is defined as:
\begin{align}\label{eq:lex}
\hspace{-.5cm} \textrm{Sem}_{Lex}(w_i,w_i') = cos(w_i,w_i')+Resnik(w_i,w_i')
\end{align}
% We compute the compatiblity score between individual words $LexMap$  using distributional and WordNet similarities. In particular, we define $LexMap(w_i,w_i') = cos(w_i,w_i')+Resnik(w_i,w_i')$, 
where $cos(w_i,w_i')$ denotes the cosine similarity between the vector space embeddings of two words $w_i$ and $w_i'$\footnote{For the ease of notation, we represent the embedding of the words with  $w_i$ as well.}, and $Resnik(w_i,w_i^{\prime})$ expresses the information content of the lowest subsumer of $\{w_i,w_i^{\prime}\}$ in WordNet. For example in Figure~\ref{fig:overview}, the semantic compatibility score incorporates lexical similarities $\mbox{Sem}_{Lex}(\mbox{dog},\mbox{ship})$, etc. 

Compatibility score between semantic relations ($\textrm{Sem}_{Pair}$) is defined by adding two components: $PairSim$ and $Analogy$ that compute how semantic relations between pairs of words are preserved in the new story:
%\begin{eqnarray*}
%  &\hspace{-.6cm}Analogy(\{w_i,w_j\}, \{w_i',w_j'\})) = \\  
%  &\textrm{analogy}(w_i,w_j ,w^{\prime}_i,w^{\prime}_j) \\
%    &+ PairSim( w_i,w_j ,w^{\prime}_i,w^{\prime}_j))
%\end{eqnarray*}
\begin{align}
PairSim = & \cos(w_i,w_j)*\cos(w^{\prime}_i,w^{\prime}_j)\label{eq:pairsim}\\
Analogy = & \cos(w_i^\prime + w_j - w_i, w_j^\prime) \label{eq:analogy}
\end{align}
$PairSim$ preserves the similarity between pairs of words $\{w_i,w_j\}$ in $s$ and the corresponding pair $\{w_i',w_j'\}$ in the new story $s^\prime$. Intuitively, if $w_i$ and $w_j$ are semantically close to each other, we would like the corresponding words to be close in the new story as well. For example in Figure~\ref{fig:overview},  `dog' and `puppy' are similar in the original story, we expect the corresponding words `ship' and `droid' to be similar in the new story. 
% We define the $PairSim$ function between pairs of content words in $s$ from the same lexical category. 
%Here, P is the set of pairs of indices of content words in $s$ from the same lexical category, a subset of CW.
The $Analogy$ function, inspired by \newcite{mikolov2013efficient}, computes the analogy of $w_j'$ from $w_i'$ given the relationship that holds between $w_i$ and $w_j$ in the vector space. For example in Figure~\ref{fig:overview}, the relation between `Sam' and `dog' is similar to the relation between `Luke Skywalker' and `ship'. 
% This calculation was introduced in \newcite{mikolov2013efficient} as demonstrating the implicit semantic regularities captured by neural word embeddings. 
% This calculation can be written as: $Analogy(w_i,w_j ,w^{\prime}_i,w^{\prime}_j)= w_j'\cdot w_i' * w_j'\cdot w_j / w_j'\cdot w_i$.
%\begin{equation}
%  \textrm{analogy}(a,b,c,d) = \nu (d) \cdot \nu (c) * \nu (d) \cdot \nu (b) / \nu (d) \cdot \nu (a)
%\end{equation}

\section{Decoding}\label{sec:decoding}

Our decoding process begins by first identifying the content words $w_i$ (nouns, verbs, adjectives and named entities) in the original problem $s$ that will be considered as \textit{initial points} for rewriting. For each of these lexical classes we extract the top-$k$ most thematic words and trivial noun compounds from the theme $t$. For example, in Figure~\ref{fig:overview}, candidate nouns are: `ships', `robots', `droids', etc., and for verbs: `blast', `soar', `command', etc. Recall that the space of candidate rewrites is large, prohibiting an exhaustive enumeration. We therefore do approximate search with a beam by considering simultaneously all possible \textit{paths} that start at the different initial points. At each step the decoder considers an additional rewrite from the list of candidates, adds it to the existing hypothesis path, and scores it according to function $\mathcal{R}$ (Equation~\ref{eq:scoring_function}). 

All the counterpart scores are locally optimal, as they factor over each new word $w_i^\prime$ or pair of $\{w_i^\prime, w_j^\prime\}$, where $w_j^\prime$ is a rewrite \textit{already} existing in the hypothesis path. At any given step we may recombine hypotheses that share the same prefix hypothesis path, and keep the top scoring one. The process terminates when there are no more rewrites left. We also experimented decoding with a variety of orderings of the text in the original problem $s$, including left-to-right, and head-first following the dependency tree of each sentence and then concatenating these linearizations; we observed that considering multiple paths achieves the best performance.

%Second, we enumerate all the possible combinations by performing greedy search with a beam, while applying local semantic, syntactic, as well as inter- and intra-sentential discourse scores at each step. %The final beam is reranked using a language model to ensure surface grammaticality. Note that our approach differs from template surface realization techniques as it does not rely on predefined choices for filling with information. It can instead arbitrarily change any part of the input text, and apply rudimentary syntactic transformations, such as reductions of noun compounds.

% The space of candidate rewrite is large, prohibiting a full ranking.
% Instead, we generate a subspace of candidates by ordering the text of $s$ and generating rewrites of increasingly longer initial segments.
% As we increase the length of the initial segment, we keep the top $N$ segments from the previous step in a beam search procedure.

\section{Data Collection}

For the set of human-authored stories $\{s\}$, we use a corpus of math word problems described in \newcite{koncel:naacl16}.
We select a subset of 150 problems targeting 5th and 6th grade levels, all of which involve a single equation in one variable.
These problems have 2.7 sentences and 29.4 words on average, 12.6 of which are considered content words by our system.
In order to tune and evaluate our model, we collect a corpus of human-authored rewrites produced by workers from Amazon Mechanical Turk based on two themes: Star Wars, and Adventure Time (a children's cartoon).

% For training and evaluation, we collect a dataset of human-authored rewrites of {500} of these problems from Amazon Mechanical Turk.
We experimented with different ways of helping to define the theme for the workers, including offering automatically generated word clouds or enforcing that a response includes one of several keywords.
In practice, we have found that using specific cultural elements as themes (such as famous movie or cartoon franchises) attracts workers who already have a strong knowledge of the theme, resulting in higher quality work.

To help explain the rewriting process, we show workers three examples of thematic rewrites with varying degrees of correlation to the original problems. 
We then show workers a random problem from the original set \{$s$\} and a corresponding equation for that problem.
We instruct the workers to ``rewrite'' the problem according to the theme, ensuring that their rewritten problem can be solved by the provided equation. 
The final dataset collection comprises of 450 human-authored rewrites. We collect 3 rewrites for 100 of the original problems for the Star Wars theme (based on the popular Star Wars sequel movies), and 3 rewrites for the rest of the 50 original problems, for the Children Cartoons Theme (\cartoon), based on the Adventure Time TV show. We keep 150 examples from the Star Wars theme for development (\swDev), and the rest 150 for testing (\swTest).

We collected the \swDev~and \cartoon~data based on workers with the ``master'' designation and at least 95\% approval rating. Then we proceeded collecting \swTest~by a subset of the authors of \swDev~who self-identify as theme experts and whose quality of work is manually confirmed. 

% The dataset is collected in four different themes; the breakdown of the number of problems collected by theme can be see in Table~\ref{dataset}.

% We gather a development dataset and in-domain test set for the Star Wars theme. \note{should specify the size of the dev set and in-domain test sets}
% To further ensure for the quality of the dataset, we collected first the development data based on workers with the ``master'' designation and at least 95\% approval rating. Then we proceeded collecting \swTest~and \cartoon~by a subset of the authors of \swDev~who self-identify as theme experts and whose quality of work is manually confirmed. 
% This produces a less noisy test dataset compared to the development set. 
% We use the development for tuning the parameters, and the rest for testing. 

%\begin{table}[t]
 % \centering
 % \begin{tabular}{|l|c|}
 %   \hline
 %   {\bf Theme} & {\bf \#} \\
 %   \hlinef
 %   Star Wars & 300\\
 %   \hline
 %   Adventure Time & 150\\
 %   \hline
    % General SciFi & 442\\
    % \hline
%    Western & 88\\
    % \hline
 % \end{tabular}
 % \caption{Data collection statistics.}
 % \label{dataset}
%\end{table}

\section{Experiments}
\subsection{Setup}

\paragraph{Implementation Details}
We pre-process the themes using the Stanford CoreNLP tools~\cite{manning-EtAl:2014:P14-5} for tokenization, Named Entity Recognition \cite{Finkel:2005:INI:1219840.1219885}, and dependency parsing \cite{chen-manning:2014:EMNLP2014}. 
For calculating salience scores, we use the ScriptBase dataset of movie scripts \cite{gorinskimovie}.
The Star Wars theme is constructed from the available script, roughly 7300 words. 
The Cartoon theme is constructed from fan-authored scripts of the first 10 episodes of the show~\cite{advtime} totaling 1370 words.
% Because our thematic options are taken from arbitrary text, we prohibit generation of offensive content by filtering $t$ to exclude offensive terms.
Since our thematic options are taken from arbitrary text, we use the lists of offensive terms published by The Racial Slur database \cite{rsdb} and FrontGate Media \cite{frontgatemedia} to filter out offensive content.
To prohibit overgeneration, we forbid the transformation of stop words or math-specific words \cite{worksheet,TACL692}.

For syntactic compatibility score Syn (Equation~\ref{eq:syn}) we use the English Fiction subset of the Google Syntactic N-grams corpus \cite{goldberg2013dataset} and train a 3-gram language model using KenLM \cite{Heafield:2011:KFS:2132960.2132986}.
For $Sem_{Lex}$, $PairSim$ and $Analogy$ (Equations~\ref{eq:lex}-\ref{eq:analogy}) we use the pretrained word embeddings of \newcite{levy2014dependency}. These embeddings are trained using dependency contexts rather than windows of adjacent words, allowing them to capture functional word similarity.
Finally, we tune the parameters of our model (Equation~\ref{eq:scoring_function}) on the development set \swDev~and pick those values\footnote{We set $\alpha=0.1$, $\beta=0.1$ and $\gamma=1$} that maximize METEOR score \cite{denkowski2014meteor} against 3 human references.
% We use the development set to tune the parameters of our model; we grid search over a set of values of the scoring function , and used  such that the METEOR score of the generated stories and human-authored rewrites are maximized. \note{Is it correct? any other details?} The best parameters for $\alpha$, $\beta$, and $\gamma$ are \note{xxx}, respectively. 

 \paragraph{Evaluation}
We compare two ablated configurations of our method against our full model (\full): \syn\ that only uses semantic and thematicity components and does not incorporate the syntactic compatibility score, \sem\ replaces the semantic coherence score with the simpler $\cos(w_i,w_i^{\prime})$, effectively rewriting only single words, and not pairs. We refrained from ablating the thematicity score as it is the core part of our model that drives the rewriting process into a new theme.

%\begin{itemize}
%\item[-syn]: Remove syntactic compatability score
%\item[-sem]: Replace semantic coherence with $\cos(w_i,w_i^{\prime})$
%\end{itemize}
We evaluate our method using an automatic metric, and via eliciting human judgments on Amazon Mechanical Turk. 
For automatic evaluation, we compute the METEOR score, comparing the output of each model for a given problem and theme to the 3 human rewrites we collected, on \swDev, \swTest~and \cartoon. METEOR is a recall-oriented metric, widely used in the MT community; the additional stemming, synonym and paraphrase matching modules make it more applicable for our use, given the nature of our rewriting task.\footnote{The average METEOR score comparing 1 annotator against the other 2 is 0.26, indicating that there are diverse “correct” strategies for solving the rewriting problem.}
 \begin{table}[t]
  \centering
  \begin{tabular}{l|c|c|c}
    \hline
    Model & \swDev & \swTest & \cartoon \\
    \hline
    \full & 31.82 &29.16 & 32.08\\
    \sem & 28.72 &25.55 & 27.55\\
    \syn & 31.92 &29.14 & 32.04\\
    \hline
  \end{tabular}
  \caption{\small METEOR results for different configuration of our model on \swDev, \swTest~and \cartoon~datasets.}
  \label{meteor}
\end{table}

For human evaluation, we conduct pairwise comparison tests, pairing \full~against a human rewrite (\human), \full~against \syn, and \full~against \sem. Participants were given a short description of the theme, and the output of each system. For each test we asked 40 subjects to select which problem they preferred over 5 pairs of outputs; we obtained a total of 200 (5x40) responses for \swTest\ and \cartoon.

In order to better understand the strengths and weaknesses of the generated stories, we conducted a more detailed human evaluation. 8 participants were presented with the output of the three automatic systems, human rewrites (\human), and a theme. The participants were asked to rate the stories across three dimensions: coherence (how coherent is the text of the problem?), solvability (can elementary school students solve it?), and thematicity (how well does the problem express them?) on a scale from 1 to 5. We collected ratings over 16 outputs from \swTest, resulting in 128 responses.
% These evaluations were collected from crowd workers with appropriate steps taken to deter spam and noise. 

\subsection{Results}

% We compare the full model \textsc{Full} described above against the two baselines \textsc{-Syn} and \textsc{-Sem}. 
% We report the average METEOR score across all problems in a dataset in .
Table~\ref{meteor} reports METEOR; we notice that removing the semantic coherence scores in \sem~hurts the performance compared to \full; this confirms our claim that semantic compatibility is crucial for building coherent stories. On the other hand, \syn\ performs similarly to \full. Closer inspection of the \syn\ system's output reveals a greater diversity in thematic elements as a result of the relaxed syntactic compatibility constraints. Hence it is more likely to have greater overlap with any of the reference rewrites, resulting in higher METEOR scores. 
% \syn\ instead mainly does more diverse but individual word changes, hence stands a better chance to agree with the varied references.

%This table shows the expected loss of across all systems when moving from the development set to the In Domain test set, although this loss is confounded by the fact that the test set authors are more expert. 
%However, the performance of the Full system by this metric is still quite reasonable. 

\begin{table}[t]
  \centering
  \begin{tabular}{l|c|c}
    \hline
    Model & \swTest & \cartoon\\
    \hline
    \textsc{FULL} & 65.0 & 57.9\\
    \textsc{  -Syn} & 35.0 & 42.1\\
    \hline 
    \hline
    \textsc{FULL} & 68.8 & 69.4\\
    \textsc{  -Sem} & 31.2 & 30.6\\
    \hline 
    \hline
    \textsc{FULL} & 17.9 & 10.0\\
    \textsc{Human} & 82.1 & 90.0\\

    \hline
  \end{tabular}
  \caption{\small Human evaluation results on pairwise comparisons between \full~and \syn, and \full~and \human, on \swTest\ and \cartoon\ datasets.}
  \label{ab}
\end{table}

\begin{table}[t]
  \centering
  \begin{tabular}{l|@{\ }c@{\ }|@{\ }c@{\ }|@{\ }c@{\ }}
    \hline
    { Model} & { Thematicity } & { Coherence } & {Solvability} \\
    \hline
    \human & 3.7 & 3.175 & 4.025\\
    \full & 3.7 & 3.025 & 3.9 \\
    \syn & 3.375 & 3.075 & 3.825 \\
    \sem & 3.325 & 2.65 & 3.7 \\
    
    \hline 
  \end{tabular}
  \caption{\small Human evaluation results for \full, \syn, \sem and \human~on thematicity, coherence and solvability on \swTest.}
  \label{likert}
  \vspace{-1ex}
\end{table}

However, a pairwise comparison between \full~and \syn\ (Table~\ref{ab}) reveals that human subjects consistently prefer the output of \full\ instead of \syn\ both for \swTest~and \cartoon. 
Table~\ref{ab} also reports that \human\ outperforms the output of the \full\ model, and 
a pairwise comparison of \full~and \sem~which yields a result in line with the METEOR scores.

%In an effort to better understand the weaknesses of our model against human-authored rewrites, we conducted a final human evaluation study to measure the quality of the output in term of more fine-grained properties. 
%Table~\ref{likert} shows that \full\ retains quite reasonable coherence, validating the importance of the syntactic and semantic components of our model. Instead it scores lower on solvability and thematicity. We hypothesize that to achieve human-level thematicity and solvability quality requires more sophisticated common-sense and domain knowledge.
% the human ratings of the generated stories and human-authored stories across three dimensions of thematicity, coherence, and solvability. The results show that the generated stories by \full are highly coherent,  quite solvable and thematic.  As expected, these ratings are higher compared to automatically generated stories.  
Table~\ref{likert} shows the results of the detailed comparison of Thematicity, Coherence, and Solvability. This table clearly shows the strong contribution of the semantic component of our system. 
The specific contribution of the syntactic component is to produce overall more solvable and thematically satisfying problems, although it can slightly affect coherence especially when automatic parses fail. 
Finally, the overall high ratings for human-authored stories across all three dimensions, confirm the high quality of the crowd-sourced stories.  

% Also the table shows comparisons between the full system and human-authored rewrites. As expected, the evaluators prefer human-authored stories. 
% Interestingly, that in about 18\%  of the cases the automatic generated stories are preferred over the human-authored stories in the scifi domain.

\begin{table}[t]
  \centering
  \begin{small}
  \begin{tabularx}{\columnwidth}{|X|}
    \hline
   {\bf Star Wars}    
   \\
 $s_1$. Wendy bought 4 new chairs and 4 new tables for her house. If she spent 6 minutes on each piece furniture putting it together, how may minutes did it take her to finish?  \\ \tabdashline
$s'_1$. {Leia bought 4 new ships and 4 new guns for her room. If she spent 6 minutes on each wasteland weapon putting it together, how many minutes did it take her to terminate?}\\ \hline 
\hline
$s_2$. My car gets 20 miles per gallon of gas. How many miles can I drive on 5 gallons of gas? \\
\tabdashline
$s'_2$. My cruiser gets 20 miles per gallon of light. How many miles can I drive on 5 gallons of light?\\
\hline
\hline
$s_3$.Tyler had 15 dogs. Each dog had 5 puppies. How many puppies does Tyler now have? \\
\tabdashline
$s'_3$. Biggs had 15 creatures. Each creature had 5 creatures. How many creatures does Biggs now have?\\
\hline
  \end{tabularx}
  \end{small}
  \caption{\small Examples of the original stories $s_i$ and rewritten math word problems $s'_i$ in Star War theme.  }
  \label{examples_star_wars}
\end{table}

\begin{table}[t]
  \centering
  \begin{small}
  \begin{tabularx}{\columnwidth}{|X|} \hline
  {\bf Cartoon} \\
$s_7$.      Dave was helping the cafeteria workers pick up lunch trays, but he could only carry 9 trays at a time. If he had to pick up 17 trays from one table and 55 trays from another, how many trips will he make? \\ \tabdashline
$s'_7$.   Finn was helping the cupboard men pick up candy bottles, but he could only carry 9 bottles at a time. If he had to pick up 17 bottles from one ring and 55 bottles from another, how many swords will he make?\\
      \hline \hline
$s_8$. If books came from all the 4 continents that Bryan had been into and he collected 122 books per continent, how many books does he have from all 4 continents combined? \\ \tabdashline
$s'_8$. If dances came from all the 4 mountains that Finn had been into and he collected 122 dances per mountain, how many dances does he have from all 4 mountains combined?\\
    \hline \hline
$s_9$. A bucket contains 3 gallons of water. If Derek adds 6.8 gallons more, how many gallons will there be in all? \\ \tabdashline
$s'_9$. A bottle makes 3 gallons of serum. If Finn adds 6.8 gallons more, how many gallons will there be in all? \\ \hline

  \end{tabularx}
  \end{small}
  \caption{\small Examples of the original stories $s_i$ and rewritten math word problems $s'_i$ in Cartoon theme. }
  \label{examples_cartoons}
\end{table}

\begin{table}[t]
  \centering
  \begin{small}
  \begin{tabularx}{\columnwidth}{|X|}
    \hline
{\bf Western} \\
$s_4$. Christian’s father and the senior ranger gathered firewood as they walked towards the lake in the park ...\\
\tabdashline
$s'_4$. Christian 's partner and the lone sheriff harvested barley as they strolled towards the hip in the orchard ... \\
\hline \hline
$s_5$. Sally had 27 cards. Dan gave her 41 new cards. Sally bought 20 cards. How many cards does Sally have now? \\
\tabdashline
$s'_5$. Madeline had 27 cigarettes. Kurt gave her 41 new cigarettes. Madeline bought 20 cigarettes. How many cigarettes does madeline have now? \\
\hline \hline
$s_6$. \small  For Halloween Megan received 11 pieces of candy from neighbors and 5 pieces from her older sister. If she only ate 8 pieces a day, how long would the candy last her? \\
\tabdashline
$s'_6$. \small For Halloween Madeline received 11 wheat of grub from proprietors and 5 wheat from her nameless partner. If she only grazed 8 wheat a day, how long would the grub last her? \\
  \hline
  \end{tabularx}
  \end{small}
  \caption{\small Examples of the original stories $s_i$ and rewritten math word problems $s'_i$ in Western theme.}
  \label{examples_western}
\end{table}

\begin{table}[t]
  \centering
  \begin{small}
  \begin{tabularx}{\columnwidth}{|X|}
    \hline
{\bf Poor Rewrites} \\
$s_{10}$. It rained 0.9 inches on Monday. On Tuesday, it rained 0.7 inches less than on Monday. How much did it rain on Tuesday?\\
\tabdashline
$s'_{10}$. It blasted 0.9 inches on Monday. On Tuesday, it blasted 0.7 inches less than on Monday. How much did it light on Tuesday?\\
\hline \hline
$s_{11}$. \small  The Ferris wheel in Paradise Park has 14 seats. Each seat can hold 6 people. How many people can ride the Ferris wheel at the same time? \\
\tabdashline
$s'_{11}$. \small The int grab in chewbacca mesa has 14 areas. Each area can hold 6 troops. How many troops can ride the int grab at the same time? \\
  \hline
  \end{tabularx}
  \end{small}
  \caption{\small Examples of the original stories $s_i$ and poorer rewrites $s'_i$ in the Star Wars theme.}
  \label{examples_poor}
\end{table}

\subsection{Qualitative Examples}
Table~\ref{examples_star_wars}--\ref{examples_western} shows some problems generated by our method. 
Recall that since our system needs no annotated thematic training data, we can easily generate from any theme where thematic text is available. 
To demonstrate this fact, we include generated examples in a Western theme from novels from the Project Gutenberg corpus. 
Many of the results of our system are very legible, with only minor agreement errors. 
Coherent, thematic semantic relations are evident in problems such as $s'_1$, where ships, guns, and weapons combine to effect the Star Wars theme; this is also evident in $s'_5$, where people with western sounding names like Kurt and Madeline trade in cigarettes, an old-fashioned pre-cursor to e-cigarettes.

In some cases, semantic inconsistencies result in weird sounding problems, such as in $s'_6$ where the main character receives ``wheat of grub''. 
But because of the syntactic compatibility component, our model scores this candidate higher because of the connection between ``wheat'' and ``graze''.

Semantic incoherence is less of a problem in the cartoon theme, where absurd interactions between characters are expected. 
However, a difficulty for our system is demonstrated in $s'_7$, where the physical entity ``swords'' is substituted for the nominalization of an event ``trips''. 
Improvements to the semantic coherence component could resolve such issues. 

Table~\ref{examples_poor} shows some instances where the rewrite algorithm produces unusable results. 
An example of under-generation is $s'_{10}$. Here, too many words are left untouched, resulting in both ungrammaticality and semantic incoherence. 
In $s'_{11}$, we witness some limitations of using word vectors. The rare word ``Ferris'' is not close to anything in the Star Wars theme, and is thus mapped almost arbitrarily to ``int'' (movie script shorthand for an interior shot). Better treatment of noun compounds and the use of phrase vectors would reduce such errors. 

%\begin{table}[t]
%  \centering
%  \begin{tabularx}{\columnwidth}{|X|}
%    \hline
%     Dave was helping the cafeteria workers pick up lunch trays, but he could only carry 9 trays at a time. If he had to pick up 17 trays from one table and 55 trays from another, how many trips will he make?\\
%  finn was helping the cupboard men pick up candy bottles , but he could only carry 9 bottles at a time . If he had to pick up 17 bottles from one ring and 55 bottles from another , how many swords will he make ?\\
%      \hline
%If books came from all the 4 continents that Bryan had been into and he collected 122 books per continent, how many books %does he have from all 4 continents combined?
%If dances came from all the 4 mountains that finn had been into and he collected 122 dances per mountain , how many dances %does he have from all 4 mountains combined ?\\
%    \hline
%A bucket contains 3 gallons of water. If Derek adds 6.8 gallons more, how many gallons will there be in all?\\
%A bottle makes 3 gallons of serum . If finn adds 6.8 gallons more , how many gallons will there be in all ?\\
%        \hline
%  \end{tabularx}
  
%  \caption{Examples of jake espisode themed math problems}
%  \label{examples}
%\end{table}

\section{Conclusion}
We formalized the problem of story rewriting as automatically changing the theme of a text without altering the underlying story and developed an approach for rewriting algebra word problems where the rewriting model optimized for a number of measures of overall text coherence. Experiments on a newly gathered dataset demonstrated our model can produce themed texts that are usually solvable.

%In future work, we plan to improve the thematicity and solvability components by incorporating domain-specific and commonsense knowledge, leveraging information extraction. We also plan to study rewriting in other domains such as children short stories and extend the model to generate math word problems directly from equations. Finally, we intend to incorporate the generated problems in educational technology and tutoring systems. 

Future work could improve the thematicity and solvability components by incorporating domain-specific and commonsense knowledge, leveraging information extraction. 
Additionally, neural network architectures (e.g., LSTMs, seq2seq) can be trained to rewrite coherently with less reliance on brittle syntactic parses. 
Additionally, we plan to study rewriting in other domains such as children's short stories and extend the model to generate math word problems directly from equations. Finally, we intend to incorporate the generated problems in educational technology and tutoring systems. 

\section*{Acknowledgments}
This research was supported by the NSF (IIS 1616112), Allen Institute for AI (66-9175), Allen Distinguished Investigator Award, DARPA (FA8750-13-2-0008) and a Google research faculty award. We thank the anonymous reviewers for their helpful comments. 

\bibliography{emnlp2016}
\bibliographystyle{emnlp2016}

\end{document}